\documentclass{INTERSPEECH2023}
\usepackage{float}
\usepackage{caption}

\interspeechcameraready

\title{The Tag-Team Approach: Leveraging CLS and Language Tagging for Enhancing Multilingual ASR}
\name{Kaousheik Jayakumar, Vrunda N. Sukhadia, A Arunkumar, S. Umesh}
\address{
  Speech Lab, Indian Institute of Technology Madras, India}
\email{kaousheik@gmail.com, ee20s008@smail.iitm.ac.in, arunkumaras10@gmail.com, umesh@ee.iitm.ac.in}

\begin{document}

\maketitle
 
\begin{abstract}

Building a multilingual Automated Speech Recognition (ASR) system in a linguistically diverse country like India can be a challenging task due to the differences in scripts and the limited availability of speech data. This problem can be solved by exploiting the fact that many of these languages are phonetically similar. These languages can be converted into a Common Label Set (CLS) by mapping similar sounds to common labels. In this paper, new approaches are explored and compared to improve the performance of CLS based multilingual ASR model. Specific language information is infused in the ASR model by giving Language ID or using CLS to Native script converter on top of the CLS Multilingual model. These methods give a significant improvement in Word Error Rate (WER) compared to the CLS baseline. These methods are further tried on out-of-distribution data to check their robustness.

\end{abstract}
\section{Introduction}
A major problem in building good ASR systems for many languages is the lack of transcribed data. To tackle the data insufficiency problem, most works have focused on pooling data from different languages to build multilingual systems. Multilingual ASR systems trained with data from many languages not only have an inherent advantage of recognizing speech corresponding to different languages used in training but also solve the data insufficiency problem by pooling. \cite{schultz1997fast} explored a common acoustic model for multiple languages based on a global phone set. Later, \cite{thomas2012multilingual} explored different training approaches using tandem and bottleneck features for multilingual Multi-layer perceptron models. In \cite{watanabe2017language}, a hybrid attention-CTC model was employed to perform grapheme-based speech recognition of 10 different languages. A similar idea of using a single multilingual end-to-end model based on attention was shown to improve performance over monolingual models for various Indian languages with the grapheme target being a union of all the characters in the languages considered \cite{8461972}. The application of adapter modules was explored to handle imbalanced datasets in a multilingual scenario in \cite{kannan2019large}. 

In multilingual systems, as the number of languages increases, the number of target units to be modeled also increases. The concept of using a common phone set, which reduces the number of target units, was earlier used in English-Hindi code switched ASR \cite{sivasankaran2018phone,dhawan2020investigating} and also in multilingual speech synthesizers across four Indian languages \cite{prakash2019building}. Previous work in multilingual ASR \cite{vishwas_icassp_CLS} proposed a Common Labeled Set (CLS) to map characters with similar sounds from different Indian languages to one common representation using their similar characteristics of Unicode representation across Indian languages. Though the CLS approach gives an improvement in performance, in order to render the decodes in their native script we require prior knowledge of the language.

The idea of adding a Language Identity (LID) token also has been investigated in many works in different contexts. Various methods have been proposed based on how the LID is incorporated in the models. \cite{9053808} proposed the use of Language ID tokens and language embedding for the low-resource multilingual ASR model to incorporate the language information in the model. {\cite{lyu2008language,mabokela2013integrated} explore having a LID predictor before feeding the identified chunks to the respective monolingual models to handle code-switch utterances. \cite{gonzalez2014real} talks about learning and predicting the LID parallelly through which the best decode is selected from different monolingual systems using the respective confidence scores. Further, integrating LID and multilingual ASR under a common framework is discussed in \cite{watanabe2017language}.

In this work, we have focused on exploring methods to improve the performance of CLS models by incorporating language information. CLS models often struggle to accurately capture the linguistic aspects of different languages, which can lead to lower accuracy. To address this challenge, we have investigated two different approaches: (i) CLS Model with CLS-to-Native Script converter [CLS2NS] and (ii) CLS model with LID token. Using CLS2NS on top of CLS-ASR model helps take care of the linguistic aspects of that language. Using CLS model with LID tokens is another way to provide language information to the model. To evaluate the efficacy of these approaches, we compared them with a conventional CLS model. Our experiments demonstrate that both the CLS model with LID and CLS model with CLS2NS approaches outperform the conventional CLS ASR model in terms of WER (word error rate). 

The rest of the paper is organized as follows. Section \ref{sec: Dataset and architecture} briefly describes the dataset and architecture details used in this work. Section \ref{sec: Baseline} discusses the Baseline Experiments. Section \ref{sec: CLS} describes the Common Label Set. Section \ref{sec: lid} details approaches we followed to use LID to build a multilingual ASR system. Section \ref{sec: Result} discusses the results of various experiments performed in this work. Section \ref{sec: Conclusion} draws important conclusions based on the results observed. 

\section{Dataset and Architecture Details}
\label{sec: Dataset and architecture}

\subsection{Dataset}
This paper utilizes publicly available labeled speech data obtained from the Universal Language Contribution API (ULCA)\footnote{\url{https://github.com/Open-Speech-EkStep/ULCA-asr-dataset-corpus}} for five Indian languages: Hindi, Gujarati, Marathi, Bengali, and Odia, which are all Indo-Aryan languages. A random subset of 200 Hours is taken as the training data for all languages. All audio files in datasets are sampled at 16kHz. The duration of training, validation, and test speech data from each language is given in Table~\ref{Dataset}

\begin{table}[th]

  \centering
\begin{tabular}{c|c|c|c}
\hline
& \textbf{Train}& \textbf{Valid} & \textbf{Test} \\
\hline \hline
\textbf{Hindi}    & 206 & 15 & 14 \\
\hline
\textbf{Marathi}  & 201 & 15 & 15 \\
\hline
\textbf{Gujarati} & 213 & 17 & 17 \\
\hline
\textbf{Bengali} & 207  & 13 & 13 \\
\hline
\textbf{Odia}   & 211 & 16 & 16 \\
 \hline
 \textbf{Total} & 1038 & 76 & 75 \\
\hline
\end{tabular}
\caption{No. of hours of data in each language}
\label{Dataset}
\vspace{-6mm}
\end{table}

\vspace{-1em}
\subsection{Architecture Details}

\subsubsection{ASR system}
\label{Architecture details}
In this paper, all ASR models are based on the Transformer framework \cite{vaswani2017attention} with joint Connectionist Temporal Classification (CTC)/attention multi-task learning with a CTC weight of 0.3 using ESPNet\cite{watanabe2018espnet}. The standard configurations for all the systems are given in Table~\ref{ASR systems}. All ASR models in this paper are trained on one A100 GPU for 50 epochs.

\begin{table}[h]
\begin{tabular}{c|c}
\hline
\textbf{Hyperparameters} &\textbf{Values} \\ \hline \hline
Feature vector dimension          & 512  \\ \hline
Number of encoder layers          & 12   \\ \hline
Encoder units                     & 2048 \\ \hline
Number of decoder layers          & 6    \\ \hline
Decoder units                     & 2048 \\ \hline
Attention heads                   & 8    \\ \hline
CTC weight                        & 0.3  \\ \hline 
\end{tabular}
\centering
\caption{Transformer model configurations}
\label{ASR systems}
\vspace{-6mm}
\end{table}




\vspace{-2em}
\section{Baseline}
\label{sec: Baseline}
\subsection{Monolingual Model}
In this setup, each language is individually modeled using a transformer model that is specifically trained on 200 hours of data from the respective language. 750 byte pair units are utilized for each individual model training. Notably, no language model is used to test these models. All other architecture details are consistent with those presented in Section \ref{Architecture details}. The results obtained from these Monolingual Models are presented in Table~\ref{tab:result}. 


\subsection{Multilingual ASR Model}
To build the Multilingual ASR model, data from all the languages are combined and used to train a single transformer model, which follows the same architecture details as presented in Section \ref{Architecture details}. Note that in this case the text is in native script of each language. The total training data amounts to 1000 hours, with 200 hours collected from each of the five languages. As language information is not explicitly provided to the model during training, the Multilingual ASR model is designed to be language-independent. The model utilizes 2500 Byte Pair Encodings (BPE) during training and does not rely on any External Language Model during testing.

\section{Common Label Set Multilingual ASR Models}
\label{sec: CLS}       

\subsection{Common Label Set}
\label{subsec: cls}

Since graphemes and phonemes in Indian languages have a strong correlation, grapheme-to-phoneme conversion can be simple and rule-based. The possibility of a reduced and universal set of target labels is afforded by the acoustic similarity between corresponding graphemes in various languages. A standard set of labels are given to phonetically similar speech sounds in Indian languages by the CLS. This work uses the CLS proposed in \cite{ramani13_ssw}. Equivalent sounds corresponding to different grapheme from different languages are given the same label. Each CLS label is made up of a string of alphabet and Roman numbers. For instance, the CLS label "aa" stands in for the sound represented by the International Phonetic Alphabet (IPA) /a:/. Table \ref{fig:cls-1} provides examples of CLS mapping with a few Indian language scripts. Unified Parser \cite{10.1007/978-3-319-45510-5_59} is used to create CLS scripts from the corresponding Native Script.  The language is determined by the Unicode range, and the language-specific rules are then applied to generate the CLS-based representation. Table \ref{fig:table_cls_lid_exp} demonstrates that the similar-sounding phones are mapped to the same CLS labels even though the scripts are different. It is important to note that reconstructing native script text from CLS text is not straightforward due to special cases such as schwa deletion, geminate correction, and syllable parsing.
\vspace{-1.5em}
\begin{figure}[h]
    \centering
    \captionsetup{justification=centering}
    \includegraphics[scale=0.6]{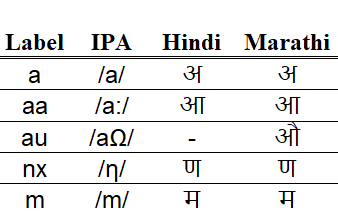}
    \captionof{table}{Examples of CLS Mapping with Native Script Characters}
    \label{fig:cls-1}
\vspace{-6mm}
\end{figure}



Using the CLS approach in Section \ref{subsec: cls}, the native script text is converted to a CLS text format for all training data from different languages. We then trained an end-to-end ASR model to predict CLS text based on speech input. However, since the eventual output needs to be in the native script, a CLS-to-Native converter model is built to convert the CLS text back to the native script text which is discussed in the subsequent sections. 


\subsection{Multilingual CLS Model}
To create the multilingual CLS model, a transformer model is utilized with 12 Encoder layers, 6 Decoder layers, and 3000 BPE units. This approach allows the model to effectively learn and identify the common sound patterns shared across multiple languages, providing a more efficient and effective means of speech recognition. By leveraging the benefits of the CLS, this model is capable of producing highly accurate and contextually relevant speech output across multiple Indian languages in the CLS target space.

\subsection{CLS-to-Native Script Converter}
\label{sec:cls-to-ns}
The CLS-to-Native Script converter [CLS2NS] is a text-to-text transformer model as shown in Figure \ref{fig:cls2ns}. In this paper, a CLS2NS is built for each language. The input to CLS2NS is CLS text and the target is the native script. We need excellent transliteration models to exploit the gains realized by the CLS system. Otherwise, the CLS2NS model might introduce errors that will affect the system’s overall performance. A transformer model with 6 encoder layers, 6 decoder layers, and 4 attention heads are used to train this model for each language. The CLS2NS models are trained using the \textit{Fairseq} toolkit \cite{ott2019fairseq}. All CLS2NS models are trained using only the text corresponding to each language's 200 hours of training data. Note that while training we use ground truth CLS and native script. As we will see, using CLS2NS provides additional gains over performance in CLS space. Additionally, the CLS2NS decoder also serves to rectify any errors that may occur in the CLS space. 
One such example is shown in Figure \ref{fig:cls2ns-error-rectify}.  
In the next section, we investigate the explicit use of language information to improve performance.
\vspace{-0.8em}
\begin{figure}[H]
    \centering
    \captionsetup{justification=centering}
    \includegraphics[scale=0.4]{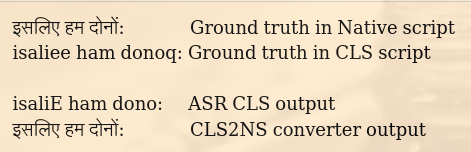}
    \caption{CLS2NS rectifies errors caused by the CLS ASR model}
    \label{fig:cls2ns-error-rectify}
\end{figure}
\vspace{-1em}
\begin{figure}[H]
    \centering
    \captionsetup{justification=centering}
    \includegraphics[width=\columnwidth]{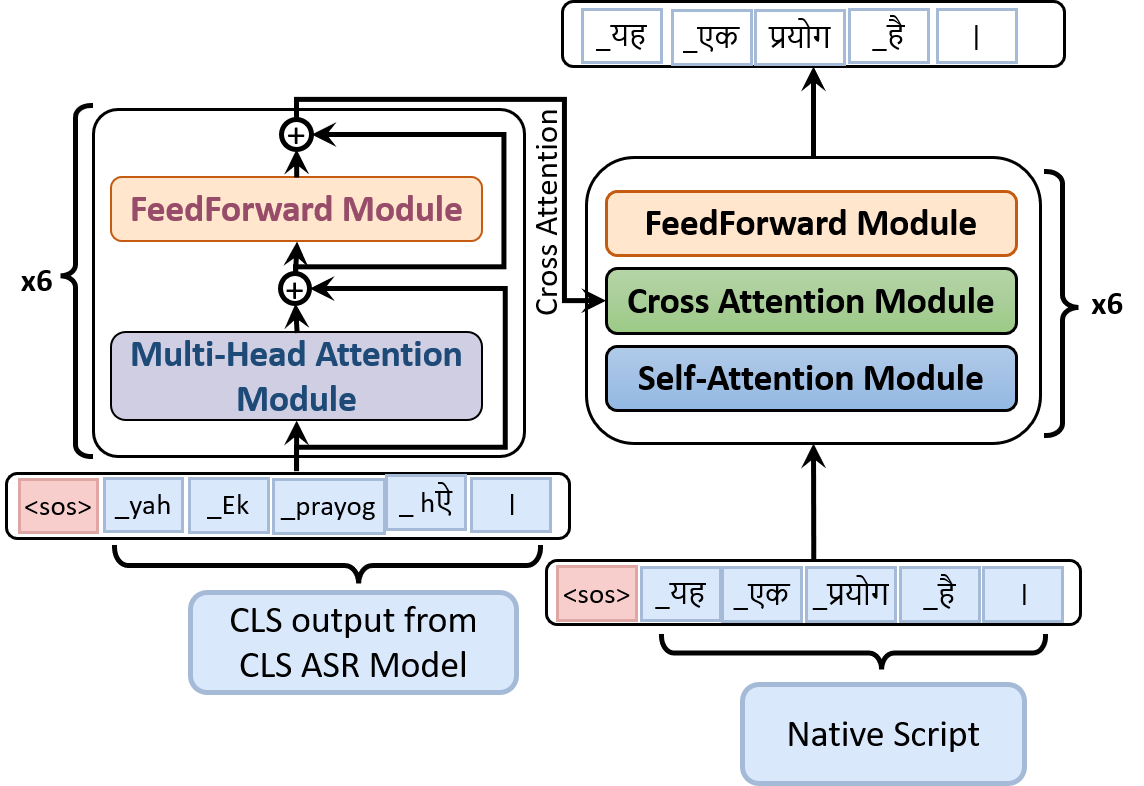}
    \caption{A Sequence-to-Sequence Transformer model is trained for CLS2NS to convert CLS text to Native Script text}
    \label{fig:cls2ns}
\end{figure}

\vspace{-2em}

\subsection{Monolingual model with Language model}
One may argue that the decoder of the CLS2NS acts like a language model for that particular language. To justify the use of the CLS2NS over simply adding an external language model in ASR model, we compared the results of the Monolingual model with the external language model and CLS ASR model with CLS2NS. In this setup training data text is only used to train the language model for each language. All Language models are based on the Transformer framework with the configuration as follows: attention units 1024, encoder units 2048, encoder layers 16, attention head 8, batchbins 350000, and learning rate 0.001 with warumplr learning rate scheduler. 
\section{Multilingual ASR Models using Language ID Token }
\label{sec: lid}
\subsection{Native Script with Language ID Multilingual ASR Models}
A different way to introduce language-specific information into an ASR model is to include \textit{language identification} (LID) tokens in the target text. This approach involves training the ASR model with the LID token, which allows the model to better distinguish between each language's unique linguistic features and sound patterns. For instance, in the LID tags, \textbf{$<$gujarati} refers to the tag being appended at the beginning and corresponds to the language Gujarati. The experiment setup for this experiment is as shown in Figure \ref{fig: lid} and embedding for LID are learnt during training. The method of using language id is shown in Table \ref{fig:table_cls_lid_exp}.

\begin{figure}[H]
    \centering
    \captionsetup{justification=centering}
    \includegraphics[width=\columnwidth]{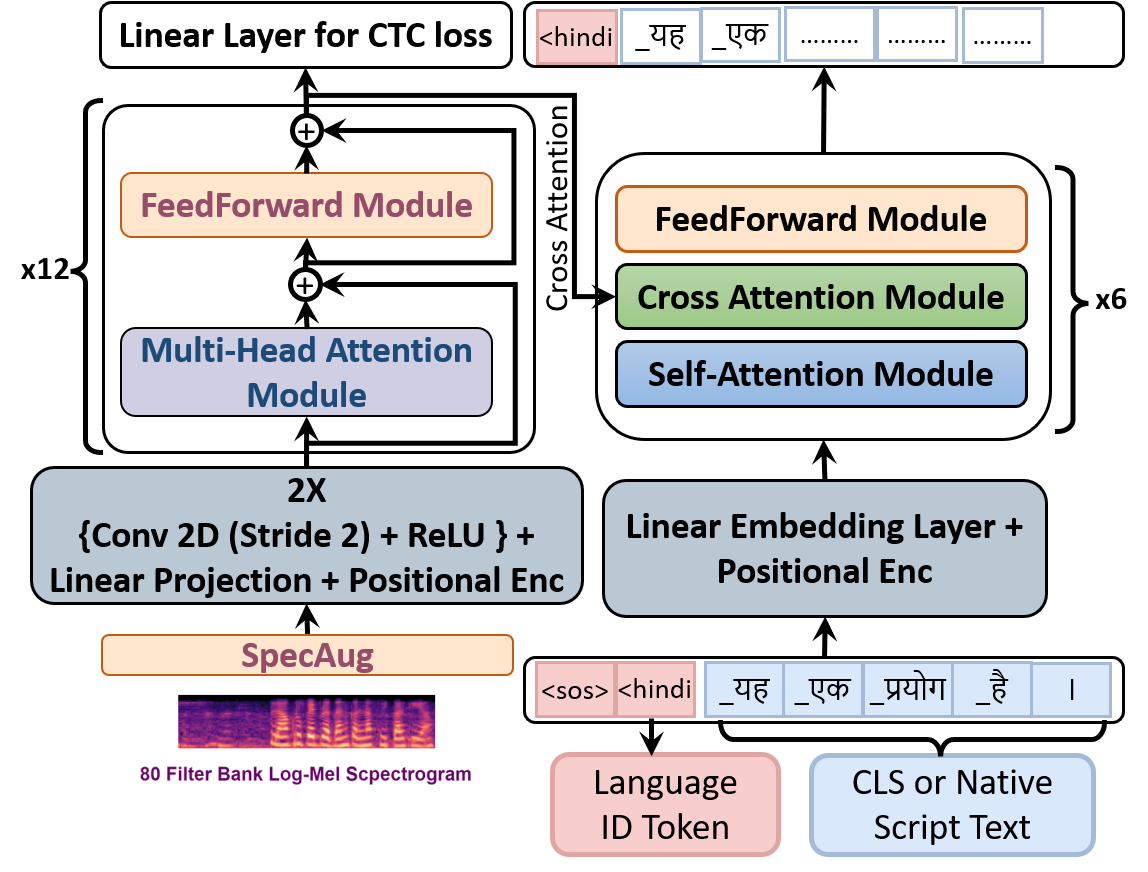}
    \caption{Experimental setup for the Multilingual ASR model with LID or CLS multilingual ASR with LID}
    \label{fig: lid}
\end{figure}

\subsection{Common Label Set with Language ID Multilingual ASR model}
A major limitation of the baseline CLS-based ASR model is that it does not have any language information while decoding the test utterance. In this study, we have investigated the use of Language Identification (LID) tokens in conjunction with CLS text during the training process. To achieve this, the transcription text was modified as illustrated in Table \ref{fig:table_cls_lid_exp}. Specifically, the CLS+LID column shows samples from the six Indian languages with corresponding LID tokens added to the beginning of the transcription text. The use of LID tokens in the decoder may be interpreted as providing means for the model to learn language aspects of that language. 

\subsubsection{CLS2NS for CLS+LID}
To convert the output of the CLS+LID model from CLS text to Native script text, mono CLS2NS or unified CLS2NS can be used. Mono CLS2NS is the same as described in section \ref{sec:cls-to-ns}. The Unified CLS2NS approach entails training the CLS2NS model using combined text data from all languages. This results in a single, comprehensive model that can be employed to seamlessly convert text from the CLS space to the native script of any language.  

\begin{figure}[]
    \centering
    \captionsetup{justification=centering}
    \includegraphics[width=\columnwidth]{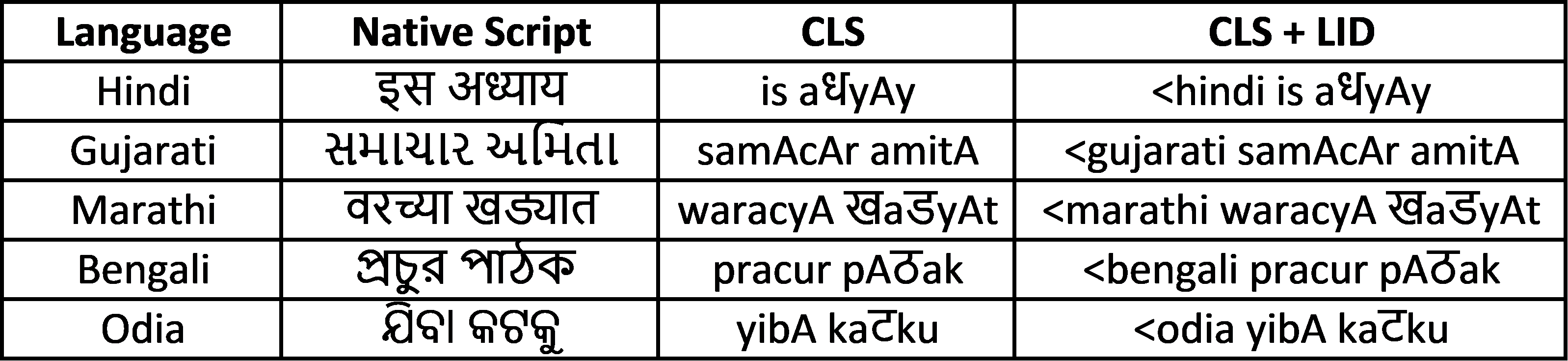}
    \captionof{table}{In multilingual the reference transcription is provided in their native scripts while in CLS they are in the common script. LID provides the language information while training }
    \label{fig:table_cls_lid_exp}
\vspace{-6mm}
\end{figure}


\begin{table*}[ht]
\centering
\captionsetup{justification=centering}
\begin{tabular}{c|c|c|c|c|c}
\hline
\textbf{Experiment}                               & \textbf{Hindi} & \textbf{Gujarati} & \textbf{Marathi} & \textbf{Bengali} & \textbf{Odia}  \\ \hline
Monolingual                                     & 17    & 26.3     & 52.2    & 23      & 32.7  \\ \hline
Monolingual + LM                                  & 16.7  & 25.7     & 51.5    & 22.3    & 32.2  \\ \hline
Multilungual Native Script                                 & 16.8  & 28.4     & 52.2    & 23.1    & 31.1  \\ \hline
Multilingual CLS                      & 16    & 27.1     & 50.9    & 22      & 29.5  \\
+ CLS2NS                        & 16    & 24.61    & 44.74   & 18.46   & 23.57 \\ \hline
Multilingual Native script with LID          & 15.4    & 24.5     & 45.2    & 20.7    & 29    \\ \hline
Multilingual CLS with LID          & \textbf{14.2}  & \textbf{22.8}     & 43.9    & 19.5    & 27    \\ 
+ CLS2NS & 15.54 & 23.4     & \textbf{43.43 }  & \textbf{18.1 }   & \textbf{23.45} \\ 
\hline
Multilingual CLS with LID + Unified CLS2NS   & 15.94 & 25.34    & 44.76   & 18.27   & 23.74 \\ \hline
\end{tabular} \\ 
\caption{WER comparison of Baseline and various Multilingual ASR model [No external LM is used except for the Monolingual + LM  experiment]}

\label{tab:result}
\vspace{-6mm}
\end{table*}
\section{Results and Discussion}
\label{sec: Result}
Results for all experiments are shown in the Table \ref{tab:result}. We begin by comparing the multilingual NS model and the multilingual CLS model with the monolingual models. We can see that the multilingual NS model is slightly better or competitive to the monolingual models. This is mainly because the encoder of the multilingual model sees more data than that of the monolingual model. But, the decoder of the multilingual model has to learn to output the text in native script of various languages and the search space of the decoder becomes large. This is the reason why it is sometimes inferior to the monolingual model. On the other hand, the multilingual CLS model, as given in row 5 of Table \ref{tab:result}, is consistently better than the monolingual models as well as the multilingual NS model. But, these results are evaluated for the CLS text as targets.

In practice, it is more useful to evaluate the performance in native script. To convert the CLS text to NS text, we train a CLS2NS as described in Section \ref{sec:cls-to-ns} for each language separately. Row 6 of Table \ref{tab:result} shows the performance of multilingual CLS model whose outputs are converted to native script by the use of CLS2NS models and it consistently outperforms the aforementioned models. We argue that the jump in performance after applying the CLS2NS is due to its language modeling capability to an extent. More specifically, we hypothesize that the decoder of the CLS2NS acts as a language model.

Past works have suggested that the use of LID tokens in multilingual systems is beneficial. We wanted to know if we can get similar benefits in multilingual CLS models by adding LID tokens to the CLS text. As we can see from Table \ref{tab:result}, adding LID tokens improves the performance of both the multilingual systems as expected. Even though the multilingual models trained with LID tokens can predict the LID during inference, for the results presented in this work, we always provide the ground truth LID since we know the language being inferenced. It is observed that these results are better than the multilingual native script with LID model. When we convert the CLS text to native script, the performance improved for three languages and degraded a little for the other two languages. In cases where the performance has improved, the improvement is not as large as compared to the non-LID counterpart. The reason for this is that adding LID to the target text already improves the language modeling power of the decoder for a given language. On top of that, the CLS2NS does not have much to offer in terms of the language model.

Finally, we wanted to see if we can unify the CLS2NS of various languages into a single model. So, we trained a unified CLS2NS which can convert the CLS text of all languages to its corresponding native script. We can see that its performance is competitive to the individual CLS2NS  with little degradation. But the advantage is that we can replace the individual models with a single unified model.

\begin{table}[H]
\centering
\captionsetup{justification=centering}
\begin{tabular}{|c|c|c|c|}

\hline
\textbf{Experiment}                                                             & \textbf{Hindi} & \textbf{Gujarati} & \textbf{Marathi} \\ \hline
Monolingual                                                            & 34.9  & 43.1     & 57.8    \\ \hline
\begin{tabular}[c]{@{}c@{}}Multilingual CLS+LID\\ +CLS2NS\end{tabular} & 31.82 & 32.04    & 37.7    \\ \hline
\end{tabular}
\caption{WER comparison for FLUERS test data [out-of-distribution]}
\label{tab: fluers}
\vspace{-6mm}
\end{table}
Additionally, we evaluated our monolingual system  on an out-of-distribution dataset. We evaluated three languages from FLEURS dataset \cite{conneau2023fleurs}. The results are shown in Table \ref{tab: fluers}. It can be clearly seen that same pattern emerges for an out-of-distribution test data also. Multilingual CLS and Multilingual CLS+LID are more robust than the monolingual systems and show up to 20.1 \% absolute improvement in WER.

Overall, we observed up to 5.55\% absolute improvement in WER compared to the multilingual with LID baseline and up to 9.25\% absolute improvement in WER compared to the monolingual baseline.
\vspace{-1em} 
\section{Conclusion}
This paper discusses the methods in which one can leverage the phonetic similarities of different languages (like Indian languages) and build a Multilingual ASR model. The data from all the languages are pooled together and phonetically similar speech sounds are assigned to Common Label Set. This reduces the number of target units for ASR. A CLS2NS is built for each language that decodes CLS text to native script text. Experiments performed  show that CLS-ASR model gives  significant improvement over the baseline monolinguistic and multilingual ASR models without CLS. Further, another method to include language-specific information to the ASR model is discussed by means of Language Identification tokens. The LID tokens are passed to the model for training in two ways: with Native Script and with CLS. To convert the output of the CLS+LID model into native script text, mono and unified CLS2NS converters are used. The experimental results show that CLS+LID model performs best among all the experiments. The Multilingual CLS and Multilingual CLS+LID models are giving the best performance also on out-of-distribution test dataset FLEURS.  

\label{sec: Conclusion}

\vspace{-1em}
\section{Acknowledgements}
 We would like to thank the Ministry of Electronics and Information
Technology (MeitY), Government of India, for providing us
with the compute resources as a part of the ”Bhashini” project.

\bibliographystyle{IEEEtran}
\bibliography{mybib,mybib_is21}

\end{document}